\documentclass{article}
\usepackage{amssymb}
\usepackage{bm}
\usepackage{amsmath}
\usepackage{booktabs}
\usepackage{multirow}
\usepackage{graphicx}
\usepackage{multicol}
\usepackage{wrapfig}
\usepackage{afterpage}
\usepackage{float}

\usepackage[preprint]{corl_2025} 

\title{Multi-critic Learning for Whole-body End-effector Twist Tracking}

\makeatletter
\newcommand{\printfnsymbol}[1]{%
  \textsuperscript{\@fnsymbol{#1}}%
}
\author{Aravind Elanjimattathil Vijayan\thanks{\fontsize{8}{8} Correspondence to \texttt{earavind@ethz.ch}.}\;\;\printfnsymbol{2},
~\textbf{Andrei Cramariuc} \printfnsymbol{2},
~\textbf{Mattia Risiglione} \printfnsymbol{2},\\
~\textbf{Christian Gehring} \printfnsymbol{4},
~\textbf{Marco Hutter} \printfnsymbol{2}\\
ETH Zurich\printfnsymbol{2}, 
ANYbotics AG\printfnsymbol{4}
}

\begin{document}
\maketitle

\begin{figure}[!tbh]
    \centering
    \includegraphics[width=1.0\linewidth]{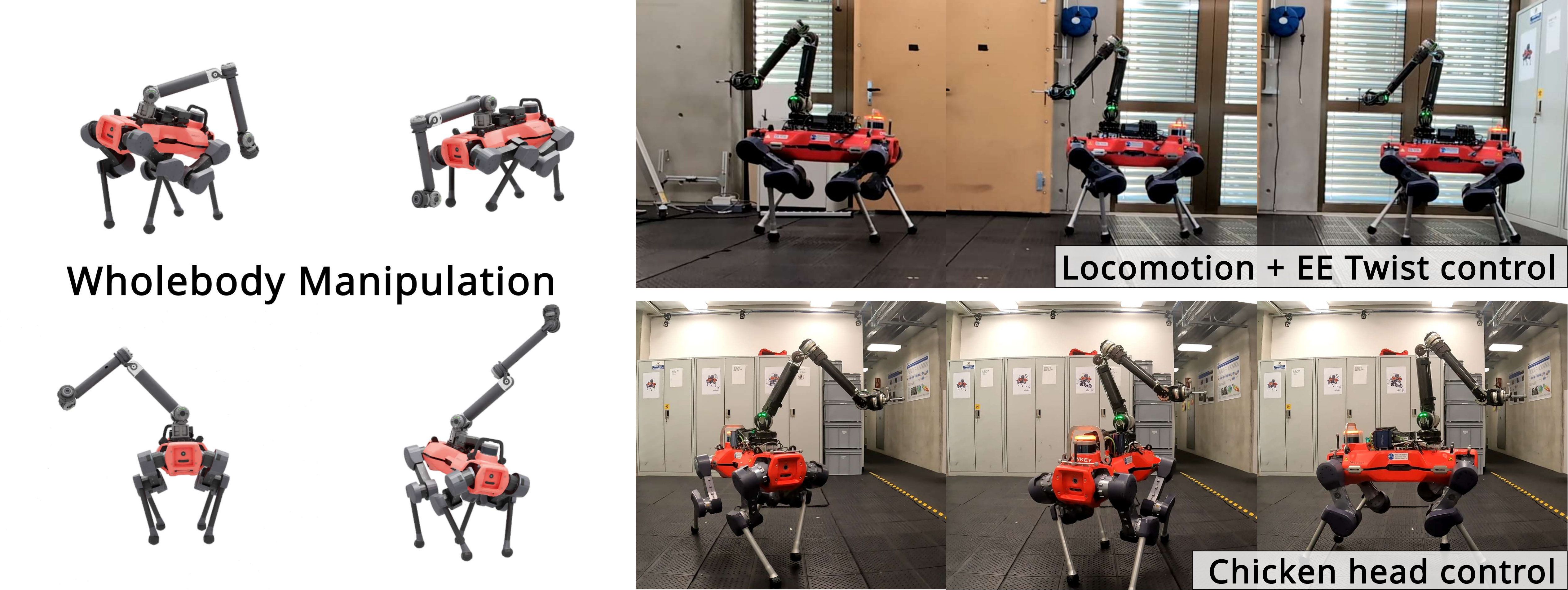}
    \caption{\textbf{Smooth and precise whole-body loco-manipulation.} We present a framework that enables quadrupedal robots to perform precise loco-manipulation through a multi-critic architecture and twist-based task space control. Our approach generates coordinated whole-body behaviors (Left) while maintaining accurate end-effector control during locomotion (Right).}
    \label{fig:title}
\end{figure}

\begin{abstract}
  Learning whole-body control for locomotion and arm motions in a single policy has challenges, as the two tasks have conflicting goals. For instance, efficient locomotion typically favors a horizontal base orientation, while end-effector tracking may require the base to tilt to extend reachability. Additionally, current Reinforcement Learning (RL) approaches using a pose-based task specification lack the ability to directly control the end-effector velocity, making smoothly executing trajectories very challenging. To address these limitations, we propose an RL-based framework that allows for dynamic, velocity-aware whole-body end-effector control. Our method introduces a multi-critic actor architecture that decouples the reward signals for locomotion and manipulation, simplifying reward tuning and allowing the policy to resolve task conflicts more effectively. Furthermore, we design a twist-based end-effector task formulation that can track both discrete poses and motion trajectories. We validate our approach through a set of simulation and hardware experiments using a quadruped robot equipped with a robotic arm. The resulting controller can simultaneously walk and move its end-effector and shows emergent whole-body behaviors, where the base assists the arm in extending the workspace, despite a lack of explicit formulations. Videos and supplementary material can be found at \href{https://multi-critic-locomanipulation.github.io/}{multi-critic-locomanipulation.github.io}.
\end{abstract}



\section{Introduction}

Robots need basic body control to perform manipulation tasks in various scenarios, such as industrial applications or service robotics.
A robust low-level whole-body controller that can track body and end-effector poses and trajectories enables the seamless integration and execution of high-level tasks~\cite{ha2024umi}. Multiple approaches to whole-body control exist, mainly divided between model-based control~\cite{sleiman2021unified, mittal2022articulated, zimmermann2021go, chiu2022collision}, reinforcement learning (RL) based methods~\cite{arm2024pedipulate, fu2022deepwholebodycontrollearning, ha2024umi, portela2025whole, wang2024arm, kaiwen2025learning, portela2024learningforcecontrollegged}, and mixes of the two~\cite{ma2022combining, liu2024visual, pan2024roboduet}. Model-based approaches have limitations and lack robustness due to the difficulty in accurately modeling interactions and the environment. RL has emerged as a staple framework for training locomotion controllers that can work in challenging environments~\cite{hwangbo2019learning, lee2020learning, miki2022learning, hoeller2024anymal}.

Whole-body control using RL faces several challenges. Firstly, locomotion and end-effector control have conflicting goals. Locomotion tends towards a horizontal body orientation for energy-efficient gaits~\cite{hwangbo2019learning, lee2020learning, miki2022learning}, but tilting the base can increase the reachability of the arm. Similarly, velocity tracking of the base is far coarser than needed to achieve precise end-effector control. Similarly, the coarse velocity tracking of the base is insufficient for precise end-effector control, leading to low accuracy in unified policies~\cite{fu2022deepwholebodycontrollearning, ha2024umi}. Solutions using separate controllers~\cite{ma2022combining} or different policies~\cite{portela2025whole} have limitations, such as requiring controller switching, not using the full workspace, or being unable to perform arm movements while in motion. A second challenge is formulating the end-effector tracking task. Current methods use goal pose tracking~\cite{liu2024visual, ha2024umi, kaiwen2025learning, zhou2020continuityrotationrepresentationsneural, portela2025whole} or force tracking~\cite{portela2024learningforcecontrollegged}. However, trajectory following using a goal pose tracking controller may not always lead to smooth end-effector control since the controller attempts to rigidly track each control point. This is a fundamental issue with such task formulation due to its lack of velocity specification.

To address the first challenge of diverging goals, we propose using a multi-critic actor architecture~\cite{mysore2022multi, cheng2023multi} that decouples the reward signals for locomotion, manipulation, and contact scheduling into individual reward groups. This simplifies the reward tuning, as separate critics allow the policy to resolve task conflicts more effectively. Furthermore, having separate critics seamlessly combines dense tracking rewards with sparse contact schedule rewards. We show that this approach is easier to use and also has desirable emergent behaviors. 
As a second contribution, we design a 6D twist-based end-effector task formulation that enables direct control over the end-effector velocity while precisely tracking both discrete poses and continuous motion trajectories. Through comprehensive simulation and hardware experiments, we validate that the twist-based formulation leads to smooth end-effector tracking during stationary operation and during dynamic locomotion of the quadrupedal robot. In summary, our contributions are:
\begin{itemize}
\setlength\itemsep{-0.2em}
    \item A multi-critic approach that trains a joint whole-body control policy that can simultaneously walk and accurately control the end-effector motion.
    \item A twist-based formulation for whole-body end-effector tracking that allows for smoothly following any end-effector trajectory at varying velocities.
    \item Experiments on simulation and hardware to demonstrate the effectiveness of our approach.
\end{itemize}


\section{Related Work}
\label{sec:related_work}
Approaches towards whole-body control can be split into two major categories: learning-based or model-based control. Model predictive controllers (MPCs) can both locomote and perform high-accuracy end-effector motions~\cite{sleiman2021unified, mittal2022articulated, zimmermann2021go}, while avoiding self-collisions~\cite{chiu2022collision}. Although these methods can be further robustified by extending the model~\cite{carron2019data} or improving the optimization framework~\cite{ferrolho2020optimizing, ferrolho2023roloma}, model-based approaches remain vulnerable to unmodeled disturbances. Especially for locomotion, reinforcement learning (RL) has emerged as the best technique to create robust and versatile controllers~\cite{hwangbo2019learning, lee2020learning, miki2022learning, hoeller2024anymal}, using large-scale parallel simulation~\cite{rudin2022learningwalkminutesusing, mittal2023orbit}.

Similarly, for end-effector control, several RL-based controllers have been proposed~\cite{ha2024umi, arm2024pedipulate, fu2022deepwholebodycontrollearning, portela2025whole, wang2024arm, kaiwen2025learning}. Some works tackle arm and base control as separate policies, where the arm is trained to reach a pose target and the base follows a separate velocity command. These methods either use RL for both~\cite{liu2024visual}, or a mix of model-based and RL control~\cite{pan2024roboduet, ma2022combining}. Such split hierarchical approaches have limitations, as the arm cannot fully leverage the base policy to extend its reachability or improve accuracy by directly commanding the leg motors.

Conflicting goals also arise when training whole-body control policies. Locomotion rewards favor horizontal base orientation for energy-efficient gaits~\cite{hwangbo2019learning, lee2020learning, miki2022learning, hoeller2024anymal}, while end-effector accuracy requires a stiff, potentially tilted base to extend arm reach. Methods balancing this trade-off~\cite{fu2022deepwholebodycontrollearning, ha2024umi} suffer from accuracy issues with tracking errors in tens of centimeters. Portela \textit{et al.}~\cite{portela2025whole} splits walking from end-effector motions but requires controller switching. In legged-wheeled systems, arm motions are more easily executed, as the presence of wheels reduces the necessity for base movements associated with walking gaits~\cite{wang2024arm, kaiwen2025learning}. Our proposed approach tackles this problem using a multi-critic approach~\cite{mysore2022multi, cheng2023multi}, enabling the locomotion and end-effector tracking problems to have separate sets of rewards, but one single policy. This emerging approach has been explored for style mixing~\cite{mysore2022multi} or locomotion~\cite{zargarbashi2024robotkeyframing, lee2024exploring}. We showcase further uses and how multi-critic RL training simplifies the tuning problem and emergently leads to desired behaviors, benefiting from accuracy and synergy between differing tasks on hardware.

Another critical aspect is target formulation for arm movement. Early works used 3D position tracking~\cite{fu2022deepwholebodycontrollearning, arm2024pedipulate}, evolving to 6D pose tracking with various representations~\cite{liu2024visual, ha2024umi, kaiwen2025learning, zhou2020continuityrotationrepresentationsneural, portela2025whole}. However, pose-based approaches struggle with smooth trajectory tracking and require high-level monitoring~\cite{ha2024umi, wang2024arm}. An alternative to pose tracking is force tracking, as demonstrated by Portela \textit{et al.}~\cite{portela2024learningforcecontrollegged}, which allows the possibility of building an impedance controller on top. In contrast, our approach circumvents these issues by directly performing twist tracking, resulting in simpler reward tuning for the entire problem. Furthermore, the twist tracking formulation provides explicit control over end-effector velocity, enabling the robot to follow complex trajectories.

\section{Method}
\label{sec:method}
\begin{figure}[!tbh]
  \vspace{-0.4cm}
    \centering
    \includegraphics[width=1.0\linewidth]{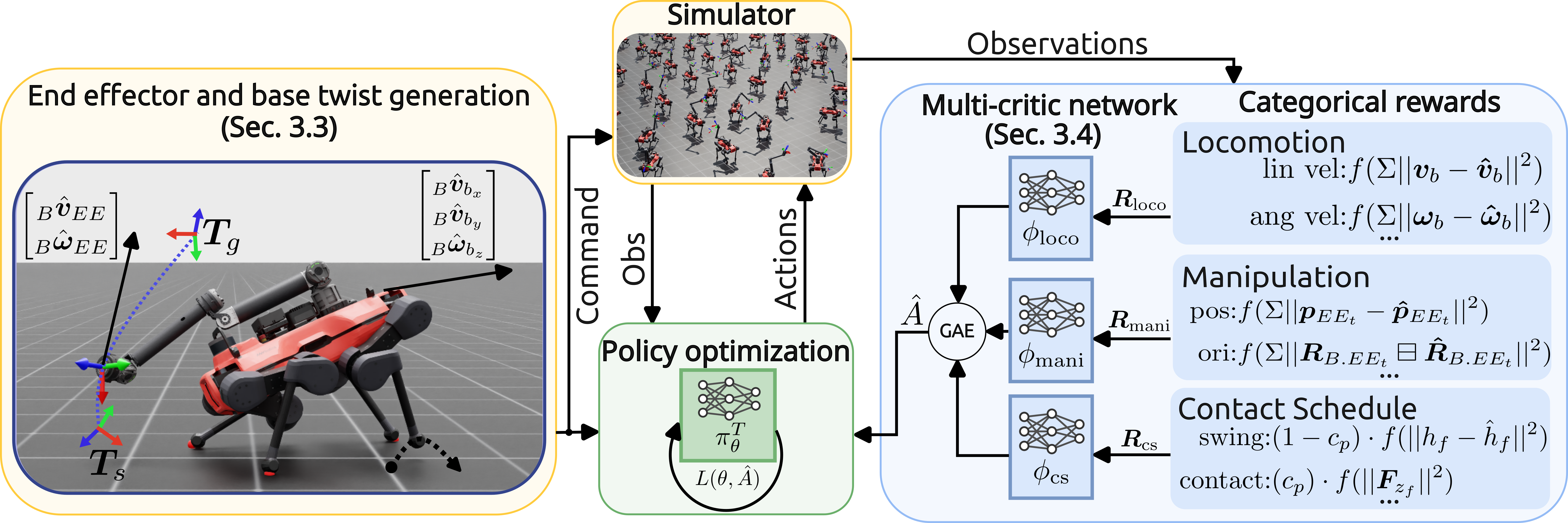}
    \caption{\textbf{Architecture for the teacher training pipeline}: Given a randomly sampled start and goal pose, the desired end-effector and base twist, along with the desired foot height, are provided as commands to the policy by the command generator. Rewards are categorically computed and consumed by separate critics, leading to individual value functions. The advantage is estimated per critic, normalized, and summed to compute the total advantage. Policy optimization is performed using Proximal Policy Optimization (PPO) to optimize the teacher policy.}
    \label{fig:enter-label}
    \vspace{-0.4cm}
\end{figure}
We aim to train a controller that takes locomotion and end-effector task commands and outputs joint position residuals, enabling coordinated whole-body motion. More specifically, we want the ability to smoothly track 6D motion trajectories with explicit control over the end-effector velocity. Lastly, we want to ensure that torso motions do not affect the manipulation task and that the robot can perform the manipulation task while stationary and during walking.

\subsection{Task definition}
\label{subsec:task_def}

We formulate the task as a Partially Observable Markov Decision Process (POMDP) and train the policy using a teacher-student architecture (Fig. \ref{fig:enter-label}) that can follow a target command $\bm{c}_t$ defined as 
\begin{equation}
    \bm{c}_t = [_B\bm{v}_b,~_B\bm{\omega}_b,~_B\bm{v}_{EE},~_B\bm{\omega}_{EE},~_B\bm{\hat{p}}_{G},~\bm{\hat{R}}_{BG},~\bm{h}_f],
\end{equation}
where $_B\bm{v}_b$ and $_B\bm{\omega}_b$ are the base linear and angular velocity targets expressed in robot's base frame, $_B\bm{v}_{EE}$ and $_B\bm{\omega}_{EE}$ are the end-effector linear and angular velocity targets expressed in robot's base frame, $_B\bm{\hat{p}}_{G}$ and $\bm{\hat{R}}_{BG}$ define the final pose in the command trajectory expressed in robot's base frame, and $\bm{h}_f$ is the desired swing heights for each foot. 

At each control time step $t$, the teacher policy $\pi_\theta^T$ generates an action $\bm{a}_t$ taking as input $\bm{s}_t~=~[\bm{h}_t,~\bm{o}_t,~\bm{p}_t,~\bm{a}_{t-1},~\bm{c}_t] \in \mathbb{R}^{187}$ consisting of the 4 step history of previous joint positions $\bm{h}_t \in \mathbb{R}^{72}$, a set of observations concerning the robot state $\bm{o}_t \in \mathbb{R}^{45}$, privilege information obtained from the simulator $\bm{p}_t \in \mathbb{R}^{32}$, the action at previous time step $\bm{a}_{t-1} \in \mathbb{R}^{18}$, and the command $\bm{c}_t \in \mathbb{R}^{20}$. For more details, please refer to the Appendix. We normalize the observations using a running buffer to ensure that the policy can learn a robust representation of the state space.

Given the observations, the policy $\pi_\theta^T$ generates an action $\bm{a}_t = [\bm{a}_t^{arm},~\bm{a}_t^{robot}]$ where $\bm{a}_t^{arm} \in \mathbb{R}^{6}$ and $\bm{a}_t^{robot} \in \mathbb{R}^{12}$ are the relative offsets to be applied. Unlike the standard approach\cite{portela2025whole} of applying actions to predefined default joint positions, we apply these offsets directly to the current joint positions. Assuming a Gaussian action distribution, the actions are normalized with a fixed standard deviation $\sigma_a$ and clamped to be within $[-1, 1]$. The joint position targets for the next time step $\bm{q}_{t+1}$ are computed as,
\begin{equation}
    \bm{q}_{t+1} = \bm{q}_t + \text{clamp}((\bm{a}_t/\sigma_a),~-1,~~1) 
\end{equation}

\subsection{Command Formulation}
\textbf{End-effector trajectory sampling}: During training, we generate manipulation trajectories by initializing the start pose $\bm{T}_s$ at the current end-effector location, while the goal pose $\bm{T}_g$ is randomly sampled using a two-stage approach. First, we sample positions within a unit sphere centered at the robot's shoulder. Any samples falling within a predefined cuboid representing the robot's torso and hip region are rejected to ensure physically feasible goals, which enables smoother training~\cite{portela2025whole}. For orientation determination, we construct a reference orientation where the x-axis aligns with gravity and the z-axis points along the vector from the torso center to the sampled position. This reference orientation is then perturbed with random rotations about each axis, bounded within $±\frac{\pi}{4}$ radians. This approach ensures that a wide range of the sampled poses are within the robot's feasible workspace. Given the start and goal pose, we implement a temporal trajectory formulation that prioritizes continuous tracking rather than terminal goal achievement. We assign a random duration to each trajectory and generate intermediate waypoints to be tracked using linear interpolation in position and spherical linear interpolation in orientation. The intermediate goals are defined as
\begin{equation}
\begin{aligned}
\bm{r}_i &= \bm{r}_s +  \alpha\cdot (\bm{r}_g - \bm{r}_s) & \bm{r} &\in \mathbb{R}^3 \\
\bm{\theta}_i &= \bm{\theta}_s \boxplus\alpha\cdot (\bm{\theta}_g \boxminus \bm{\theta}_s) & \bm{\theta} &\in SO(3) \\
\end{aligned}
\end{equation}
where $\bm{r}_i$ and $\bm{\theta}_i$ represent the intermediate position and orientation, $\bm{r}_s$ and $\bm{\theta}_s$ represent the start position and orientation, and $\bm{r}_g$ and $\bm{\theta}_g$ represent the final goal position and orientation, and $\alpha$ is the interpolation factor. The operators $\boxplus$ and $\boxminus$ denote addition and subtraction in the Lie group $SO(3)$.

\textbf{End-effector command}: While most pose tracking methods in the literature~\cite{liu2024visual, ha2024umi, kaiwen2025learning, zhou2020continuityrotationrepresentationsneural, portela2025whole} define the task to track a final goal pose within a given time, our task formulation is a continuous trajectory tracking problem with explicit velocity control. Hierarchical approaches, which send intermediate goal poses to a pose-based tracking controller to achieve trajectories, lack critical information about the required end-effector velocity. This deficiency leads to jerky end-effector motions as the policy attempts to rigidly maintain each intermediate pose without smoothly transitioning between them. To address these limitations, we introduce an end-effector twist-based command formulation that explicitly incorporates velocity information. More specifically, we define the command as:
\begin{equation}
\bm{v}_{EE} = \frac{1}{\Delta t} (\bm{r}_i - \bm{r}_{EE})\ \ \text{and}\ \ \bm{\omega}_{EE} = \frac{1}{\Delta t} (\bm{\theta}_i \boxminus \bm{\theta}_{EE})
\end{equation}
where, $\bm{r}_{EE}$ and $\bm{\theta}_{EE}$ are the current end-effector position and orientation respectively, and $\Delta t$ is the control time step.
Finally, the end-effector command to the policy is defined as the stacked vector of $\bm{v}_{EE}$, $\bm{\omega}_{EE}$, and the final goal pose $\bm{T}_g$ along the entire trajectory.

We observe that training exclusively with global feed-forward trajectories as defined above can lead to a sparse distribution of positive tracking examples, leading to extended training times before the policy grasps the task. To mitigate this, we define a subset of the trajectories locally. A local trajectory is created by re-initializing the start pose of the trajectory at every control time step. This approach ensures that the policy frequently encounters states near the desired trajectory, providing a denser representation of the reward landscape and accelerating the learning process. Similarly to \cite{portela2025whole}, we represent the end-effector trajectory in a robot-centric task frame that is agnostic to the torso height, pitch, and roll orientation. This choice decouples the task from the torso motion, allowing the policy to learn a coordinated whole-body motion to satisfy a precise end-effector motion.

\textbf{Foot trajectory generation}:  computes the desired foot heights above the ground during locomotion. A gait phase $\phi$ and foot phase offsets $\bm{\theta} = [\theta_{LF},~\theta_{RF},~\theta_{LH},~\theta_{RH}]$ are used to generate a sinusoidal swing foot trajectory pattern~\cite{pan2024roboduet}, where the subscripts denote the four legs: Left Front (LF), Right Front (RF), Left Hind (LH), and Right Hind (RH) . The height of the feet is computed as:
\begin{equation}
\begin{split}
\bm{h}_f &= h_{max}[\text{sin}(2\pi \theta_{LF}),~\text{sin}(2\pi \theta_{RF}),~\text{sin}(2\pi \theta_{LH}),~\text{sin}(2\pi \theta_{RH})] \\
\bm{\theta}  &= [\phi+\theta_{LF},~\phi + \theta_{RF},~\phi + \theta_{LH},~\phi + \theta_{RH}]
\end{split}
\end{equation}
During an episode, the gait phase is updated according to a fixed gait frequency, and the foot phase offsets are fixed to correspond to a static gait pattern. 

\textbf{The base command}: consists of the linear velocity of the torso $_B\bm{v}_b = [_B\bm{v}_{b_x}, _B\bm{v}_{b_y}]$ and the angular velocity $_B\bm{\omega}_b$ along the Z-axis wrt. the base frame. During training, this velocity command is periodically resampled during the episode from a uniform distribution within the range $[-0.25, 0.25]$.

\subsection{Multi-critic Actor Learning}
\label{subsec:mcal}
Multi-critic actor learning has previously been shown to be effective in multi-task RL problems~\cite{mysore2022multi} and to effectively manage mixtures of dense and sparse rewards~\cite{zargarbashi2024robotkeyframing}. Fu \textit{et al.}~\cite{fu2022deepwholebodycontrollearning} introduced two critic networks and used curriculum learning to gradually mix two advantage functions to combine the learning of arm and leg actions in a unified policy. In this work, we categorize rewards into three distinct groups: locomotion, manipulation, and foot contact schedule. For each group, we define a dedicated critic $\phi_i$ that estimates the corresponding value function $V_{\phi_i}(.)$. The overall advantage $\hat{A}$ is calculated as the sum of the normalized advantage estimates $\hat{A}_i$ from each critic. Following the PPO framework, we compute the policy surrogate loss as
$$
L(\theta) = \hat{\mathbb{E}}[min(r_t(\theta)\hat{A},~clip(r_t(\theta),~1-\epsilon,~1+\epsilon)\hat{A})]
$$
where $r_t(\theta)$ is the ratio of new to the old policy probabilities, $\epsilon$ is the PPO clipping parameter, and $\hat{\mathbb{E}}$ is the empirical average over the batch.
The three reward groups are structured as follows:
\begin{itemize}
    \vspace{-2mm}
    \setlength\itemsep{-0.2em}
    \item \textbf{Locomotion}: base velocity tracking rewards and regularization terms.
    \item \textbf{Manipulation}: end-effector pose tracking rewards  and regularization terms.
    \item \textbf{Contact schedule}: rewards and regularization terms for tracking foot contact states.
\vspace{-2mm}
\end{itemize}
Notably, we introduced the contact schedule as a separate group due to the low-frequency nature of gait patterns, which results in a relatively sparse reward distribution compared to the other groups.

\section{Results}
\label{sec:result}

We deploy our controller on the ANYmal D robot~\cite{hutter2016anymal} from ANYbotics AG mounted with a Dynaarm manipulator from Duatic AG. The robot has a total of 18 actuators, of which 12 actuate the legs. While ANYmal's motion controller is executed at 400 Hz, our control policy is executed with a decimation rate of 8:1, resulting in 50 Hz on the onboard computer~\cite{portela2025whole, miki2022learning, 10610271}.

\begin{figure}[!tbh]
        \centering
        \includegraphics[width=1.0\linewidth]{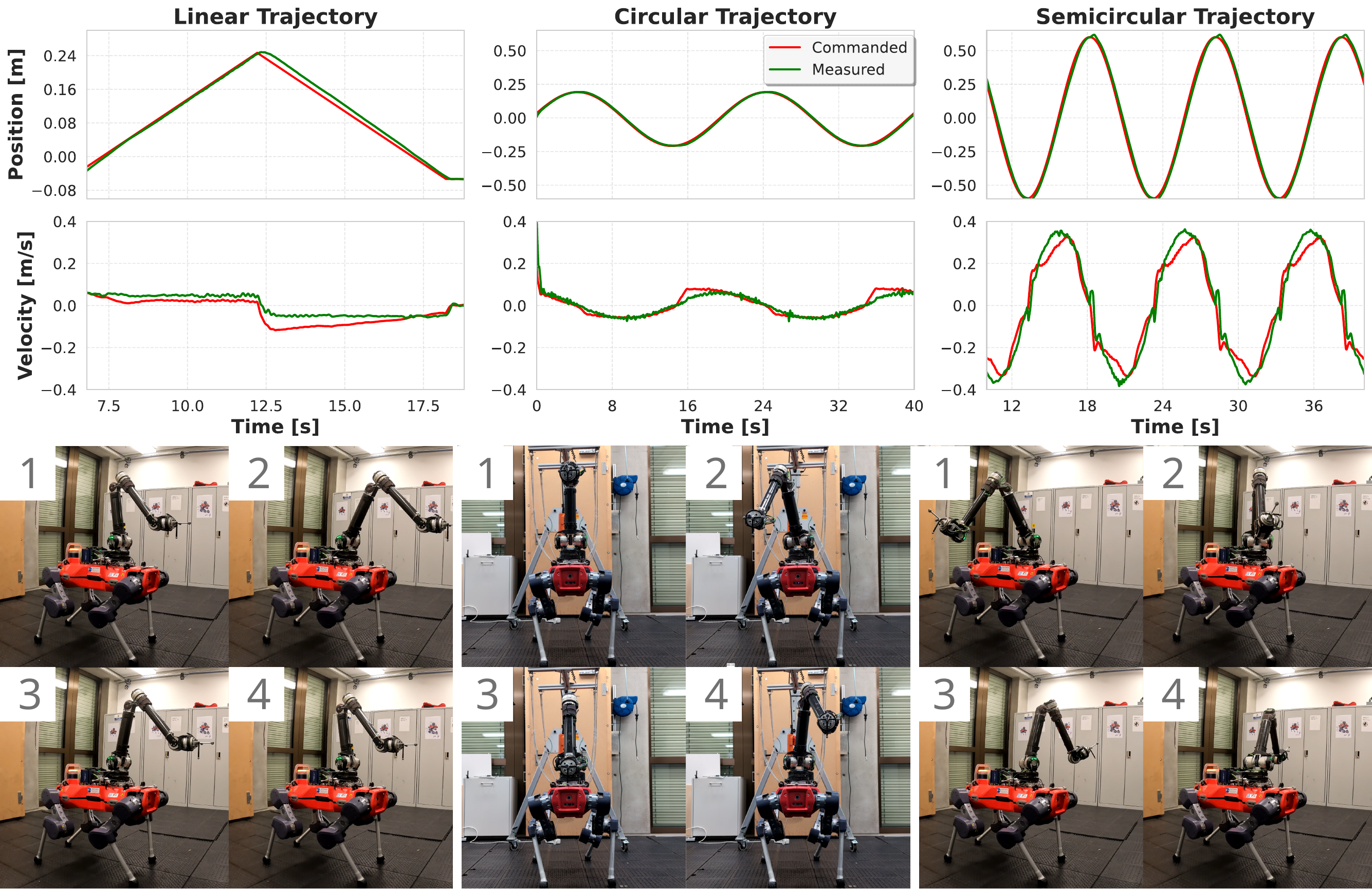}
        \caption{\textbf{End-Effector Tracking Performance for Different Trajectories}: Desired and measured position and velocity for tracking linear, circular, and semicircular trajectories (left to right).}
        \label{fig:xyz_exp_tracking}
    \end{figure}
\textbf{Tracking performance of different trajectory types on hardware}: To assess the tracking performance of the policy, we evaluate its ability to track the following trajectory types:
\vspace{-2mm}
\begin{itemize}
\setlength\itemsep{-0.2em}
    \item \textbf{Straight line}: The policy is commanded to move 30 cm along each axis.
    \item \textbf{Circle}: The policy is commanded to move in a circle on the YZ plane with a radius of 20 cm at a fixed distance in front of the robot.
    \item \textbf{Semicircle around robot}: To evaluate trajectories over an extended workspace, the policy is commanded a trajectory of radius 60 cm radius around the robot.
  \end{itemize}
  \vspace{-2mm}
  During these experiments, the end effector orientation is actively controlled to point outwards along the vector from the torso to the end effector while aligning the X-axis of the end-effector frame with gravity vector. In Fig. \ref{fig:xyz_exp_tracking}, we show the tracking performance for all three trajectory types.

\textbf{Tracking trajectories with varying velocities on hardware}:

\begin{wraptable}{l}{0.35\textwidth}
\vspace{-4mm}
\centering
\resizebox{0.35\textwidth}{!}{
\begin{tabular}{l c c c}
\toprule
\textbf{Motion} & $\hat{\bm{v}}_{EE}$ & $\delta \bm{r}$ &$\delta \dot{\bm{r}}$ \\
\midrule
\multirow{2}{*}{Linear} 
    & 0.05 & 0.0166 & 0.0519 \\
    & 0.1  & 0.0175 & 0.0623 \\
\midrule
\multirow{3}{*}{Circular} 
    & 0.05 & 0.0220 & 0.0609 \\
    & 0.1  & 0.0274 & 0.0898 \\
    & 0.2  & 0.0348 & 0.1781 \\
\midrule
\multirow{2}{*}{Semicircular}
    & 0.1 & 0.0480 & 0.2339 \\
    & 0.2 & 0.0732 & 0.3590 \\
\bottomrule
\end{tabular}
}
\caption{Tracking performance in end-effector position error (L2 norm) $\delta \bm{r}$ (m) and velocity error $\delta \dot{\bm{r}}$ (m/s) for robot linear motions XYZ executed on hardware at specified distances and speeds $\hat{\bm{v}}_{EE}$ (m/s). }
\label{tab:tracking_performance}
\vspace{-6mm}
\end{wraptable}
To demonstrate the capability of the controller to execute trajectories at varying velocities, we execute the three trajectory types described above at different commanded velocities. The RMSE end-effector tracking error in position and velocity while tracking trajectories at different speeds is listed in Table \ref{tab:tracking_performance}. As also shown in Fig. \ref{fig:xyz_exp_tracking}, the controller achieves good velocity tracking across different velocity ranges. Although the accuracy is lower at slower speeds, particularly for linear trajectories, the policy maintains low position errors throughout the execution. It must be noted that circular and semicircular trajectories being evaluated are not part of the training distribution, since the trajectories in training are always computed between start and goal pose through linear interpolation.

\begin{figure}[!tbh]
\centering
    \includegraphics[width=1.0\linewidth]{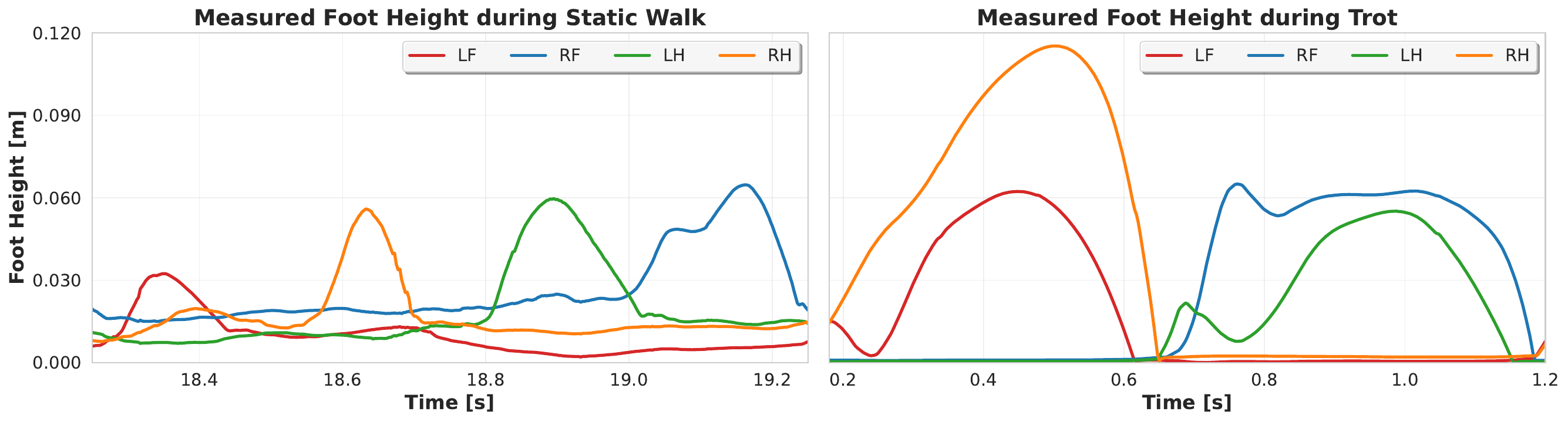}
    \caption{\textbf{Contact Schedule Adaptation to Unseen Gait Patterns}: Measured height of Left Fore (LF), Right Fore (RF), Left Hind (LH) and Right Hind (RH) feet for (a) static walking measured on hardware, and (b) trot gait measured in simulation.}
    \label{fig:foot_tracking}
\vspace{-8mm}
\end{figure}

\textbf{Effect of multi-critic architecture on gait tracking in simulation}:
Multi-critic architectures have previously demonstrated effectiveness in multi-task RL problems. While the agent is trained only with a static walking gait pattern, we observe that the policy generalizes to perform a trotting gait at runtime by simply modifying the commanded foot swing trajectories. The measured foot trajectories for both static walk and trotting gaits are shown in Fig. \ref{fig:foot_tracking}.
This result demonstrates that separating the contact schedule into a separate critic enables the policy to learn a more general representation of gait patterns, allowing it to adapt foot timings based on the input command. We note that such generalization has not been previously reported in the literature, highlighting the efficacy of our multi-critic architecture.

\textbf{Comparison with existing methods in simulation}:
To evaluate our proposed approach against existing methods, we conduct comparative experiments using a semicircular end-effector trajectory, both while the robot is stationary and during locomotion. We integrate our arm control with three different locomotion policies: Ma~\cite{ma2022combining}, Taka 3-DoF~\cite{miki2022learning}, and Taka 6-DoF~\cite{10610271}. 

\begin{figure}[!tbh]
\centering
\begin{minipage}[t]{0.38\textwidth}
\centering
\vspace{-6mm}
\begin{table}[H]
\centering
\resizebox{\textwidth}{!}{%
\begin{tabular}{l l c c}
\toprule
\textbf{State} & \textbf{Controller} & $\delta \bm{r}$ [m] &$\delta \bm{\theta}$ [$^\circ$] \\
\midrule
\multirow{4}{*}{Standing} 
    & Proposed & \textbf{0.0176} & 1.815 \\
    & Taka 3-DoF \cite{miki2022learning} & 0.0580 & 4.196 \\
    & Taka 6-DoF \cite{10610271} & 0.0439 & 3.608 \\
    & Ma \cite{ma2022combining} & 0.0386 & 3.515 \\
    & Portela \cite{portela2025whole} & 0.0391 & \textbf{1.756} \\
\midrule
\multirow{4}{*}{Walking}
    & Proposed & \textbf{0.0358} & \textbf{2.872} \\
    & Taka 3-DoF \cite{miki2022learning} & 0.4180 & 4.711 \\
    & Taka 6-DoF \cite{10610271} & 0.0436 & 4.974 \\
    & Ma \cite{ma2022combining} & 0.0483 & 6.432 \\
    & Portela \cite{portela2025whole} & - & - \\
\bottomrule
\end{tabular}}
\caption{Average end-effector tracking errors: $\delta \bm{r}$ (mean Euclidian norm) for position and $\delta \bm{\theta}$ (mean angular error) for orientation in simulation.}
\label{tab:controller_comparison}
\end{table}
\end{minipage}
\hfill
\begin{minipage}[t]{0.58\textwidth}
\centering
\vspace{-4mm}
\includegraphics[width=\textwidth]{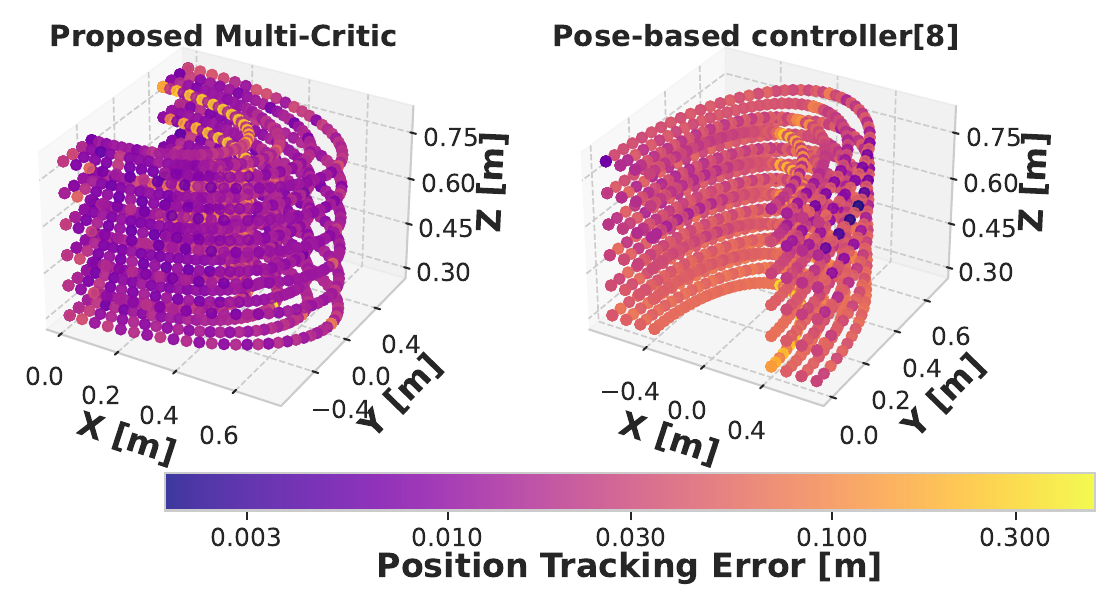}
\caption{Workspace sweep trajectory wrt. base used for evaluating tracking performance in simulation.}
\label{fig:workspace_sweep}
\end{minipage}
\vspace{-6mm}
\end{figure}
To achieve this, we apply the leg actions generated by the respective locomotion policy while executing the arm actions produced by our proposed approach. We compute the tracking end effector tracking errors using the measurements obtained from the state estimator. As shown in Table \ref{tab:controller_comparison}, while all combinations achieve satisfactory end-effector tracking, our proposed whole-body locomotion policy demonstrates superior performance with the lowest tracking errors.

For comparison with existing pose-based controllers, we evaluate against the whole-body policy from Portela et al.~\cite{portela2025whole}. Since their policy is not designed for locomotion, we present results only for stationary experiments. Our approach achieves significantly better task space trajectory tracking in position while maintaining comparable orientation accuracy. Notably, stationary policies typically have a substantial advantage during training due to enhanced stability from continuous leg contact. This makes our results particularly significant, as our proposed approach delivers superior performance while being able to perform loco-manipulation.

In addition to the linear and circular trajectories introduced above, we introduce a workspace sweep trajectory that extends the semicircular trajectory by sweeping across various radii and heights. As in Fig. \ref{fig:workspace_sweep}, our proposed approach demonstrates improved end-effector tracking performance across the evaluated workspace compared to a pose-based alternative. In Table \ref{tab:rebuttal_controller_comparison}, we present an extended comparison of end-effector tracking performance for linear, circular, and workspace sweep trajectories, and provide results also against a whole-body MPC controller~\cite{sleiman2021unified}. Our proposed approach consistently outperforms all other methods across all trajectory types and both standing and walking scenarios, achieving the lowest position and orientation errors as well as velocity tracking errors, demonstrating whole-body loco-manipulation capabilities.
\begin{table}[h]                           
 \centering
\resizebox{1.0\textwidth}{!}{%
\begin{tabular}{|l|l|c|c|c|c|c|c|c|c|}
\hline
\multirow{2}{*}{\textbf{Type}} & \multirow{2}{*}{\textbf{Controller}} & \multicolumn{4}{c|}{\textbf{Standing}} & \multicolumn{4}{c|}{\textbf{Walking}} \\
\cline{3-10}
 &  & $\delta \bm{r}$ [m] & $ \delta \bm{\theta}$ [$^\circ$] & $\delta \bm{\dot{r}}$ [m/s] & $\delta \bm{\dot{\theta}}$ [rad/s] & $\delta \bm{r}$ [m] & $ \delta \bm{\theta}$ [$^\circ$] & $\delta \bm{\dot{r}}$ [m/s] & $\delta \bm{\dot{\theta}}$ [rad/s] \\
\hline
\multirow{6}{*}{Linear} & \textbf{Multi-critic (Ours)} & \textbf{0.0095} & \textbf{1.6084} & \textbf{0.0568} & \textbf{0.1756} & \textbf{0.0093} & \textbf{1.9926} & \textbf{0.0559} & \textbf{0.2176} \\
 & Taka 3-DoF [17] & 0.0474 & 10.1269 & 0.2934 & 1.1056 & 0.0408 & 16.5597 & 0.2514 & 1.8080 \\
 & Taka 6-DoF [30] & 0.0271 & 8.2354 & 0.1648 & 0.8983 & 0.0370 & 9.9586 & 0.2285 & 1.0868 \\
 & Ma [12] & 0.0232 & 8.0863 & 0.1411 & 0.8823 & 0.0573 & 14.2652 & 0.3562 & 1.5563 \\
 & Portela [8] & 0.0240 & 1.7285 & 0.1421 & 0.7562 & - & - & - & - \\
 & Whole-body MPC & 0.0207 & 3.9522 & 0.1291 & 0.4311 & 0.0195 & 2.8417 & 0.1221 & 0.3100 \\
\hline
\multirow{6}{*}{Circular} & \textbf{Multi-critic (Ours)} & \textbf{0.0080} & \textbf{0.9338} & \textbf{0.0506} & \textbf{0.1018} & \textbf{0.0137} & \textbf{1.5921} & \textbf{0.0865} & \textbf{0.1736} \\
 & Taka 3-DoF [17] & 0.0446 & 12.3341 & 0.2774 & 1.3449 & 0.0259 & 11.8966 & 0.1611 & 1.2986 \\
 & Taka 6-DoF [30] & 0.0234 & 9.7642 & 0.1447 & 1.0658 & 0.0460 & 40.1593 & 0.2865 & 4.3794 \\
 & Ma [12] & 0.0269 & 36.6286 & 0.1663 & 3.9952 & 0.0425 & 7.8882 & 0.2653 & 0.8600 \\
 & Portela [8] & 0.0399 & 2.7328 & 0.2358 & 1.1981 & - & - & - & - \\
 & Whole-body MPC & 0.0239 & 3.2007 & 0.1497 & 0.3491 & 0.0327 & 4.2238 & 0.2044 & 0.4607 \\
\hline
\multirow{6}{*}{\shortstack{Workspace\\sweep}} & \textbf{Multi-critic (Ours)} & \textbf{0.0233} & \textbf{3.9257} & \textbf{0.1477} & \textbf{0.4323} & \textbf{0.0408} & \textbf{8.1143} & \textbf{0.2520} & \textbf{0.8876} \\
 & Taka 3-DoF [17] & 0.1415 & 13.6202 & 0.8812 & 1.4878 & 0.0858 & 9.2738 & 0.5329 & 1.0137 \\
 & Taka 6-DoF [30] & 0.0821 & 7.6319 & 0.5094 & 0.5094 & 0.0900 & 9.8273 & 0.5591 & 1.0736 \\
 & Ma [12] & 0.0906 & 8.6375 & 0.5626 & 0.9454 & 0.1514 & 16.8287 & 0.9439 & 1.8366 \\
 & Portela [8] & 0.0486 & 2.1633 & 0.2804 & 0.9737 & - & - & - & - \\
 & Whole-body MPC & 0.3244 & 38.2293 & 2.0277 & 4.1702 & 0.3346 & 39.5757 & 2.0915 & 4.3170 \\
\hline
\end{tabular}%
}
\caption{Extended comparison with trajectory and velocity tracking errors in simulation.}
\label{tab:rebuttal_controller_comparison}
\vspace{-8mm}
\end{table}

\begin{figure}[!tbh]
\centering
    \includegraphics[width=1.0\linewidth]{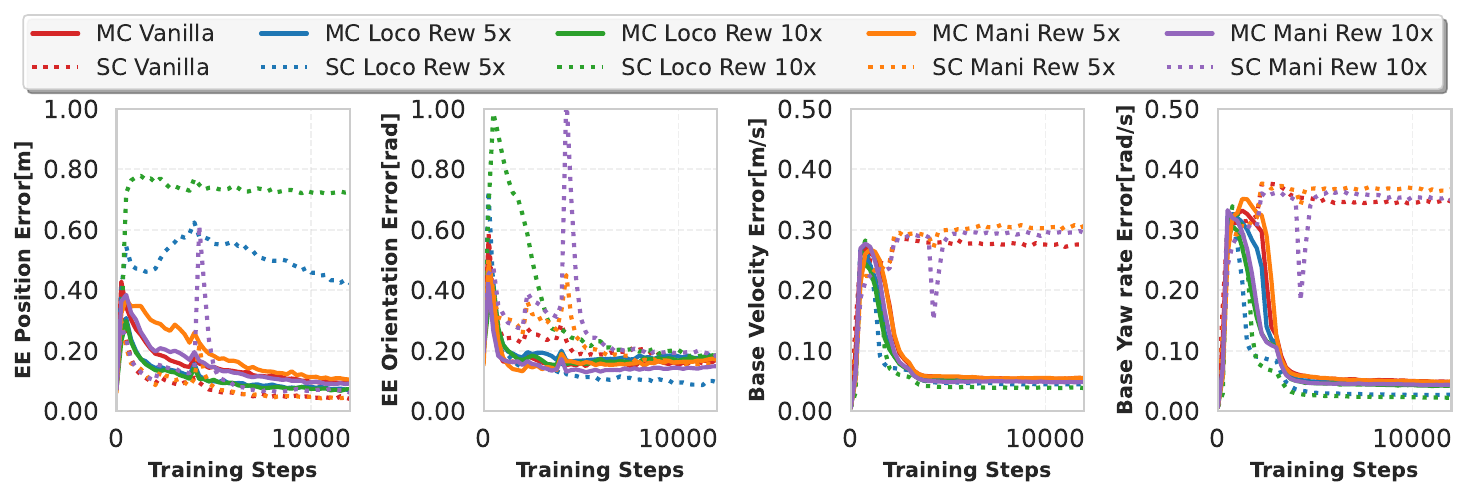}
   \vspace{-4mm}
    \caption{\textbf{Reward sensitivity analysis of single-critic vs multi-critic frameworks}: Multi-critic policy is robust against reward scaling across categories whereas the single-critic policy only learns to perform the objective with higher reward scale.}
    \label{fig:reward_sensitivity_analysis}
\vspace{-4mm}
\end{figure}
\textbf{Reward Sensitivity Analysis - Single-Critic vs Multi-Critic Learning}:
Since our approach splits rewards categorically into locomotion, manipulation, etc., we multiply all rewards within each category by a fixed multiplicative factor and evaluate the resulting policy performance for reward sensitivity analysis. Specifically, we conduct experiments scaling locomotion and manipulation rewards by factors of 5 and 10. 
As shown in Fig.~\ref{fig:reward_sensitivity_analysis}, the multi-critic approach remains insensitive to these weighting factors, while the single-critic approach only learns to perform either locomotion or manipulation and not to perform both simultaneously. When scaling the locomotion reward in the single-critic approach, we observe improved locomotion performance but deteriorated end-effector tracking. In contrast, scaling the manipulation reward improves the end-effector tracking performance but degrades locomotion quality. These experiments demonstrate that the proposed multi-critic approach is robust to variations in relative weightage between reward categories, making it effective for combining multiple learning objectives without careful reward tuning.


\section{Conclusion}
\label{sec:conclusion}
In this work, we presented a whole-body loco-manipulation framework for quadrupedal robots that enables continuous and smooth trajectory tracking while stationary and during locomotion. Through extensive simulations and hardware experiments, we demonstrated the effectiveness of our twist-based command formulation in achieving precise end-effector velocity control. Our novel multi-critic architecture successfully combines locomotion, manipulation, and contact scheduling without requiring complex training curricula or reward tuning. Comparative evaluations against existing non-whole-body and pose-based controllers validate the superior performance of our approach.

\section{Limitations}
The current implementation has two main limitations. First, locomotion performance is restricted to moderately rough terrains, as robust rough-terrain controllers were not the focus of this work. In future work, we plan to address this by including reward formulations and perception mechanisms from existing rough-terrain locomotion frameworks. Our multi-critic formulation should greatly simplify the tuning involved in incorporating these new terms. However, having to train the whole-body teacher to incorporate both robust arm maneuvers and perceptive locomotion will significantly increase training times and make later tuning of individual behaviors more difficult. The second limitation is that during whole-body maneuvers, the robot occasionally reaches the limits of the shoulder joint, triggering safety mechanisms. In the future, this can be addressed by incorporating joint limits during policy training, either as penalties, terminations, or, for example, through the use of constrained PPO. 


\acknowledgments{We thank the reviewers of CoRL 2025 for their constructive and insightful feedback. This research was supported by ANYbotics AG and the Swiss National Science Foundation through the National Centre of Competence in Automation (NCCR automation). This project also received funding under the European Union’s Horizon Europe Framework Programme under grant agreement No 101121321.
}


\bibliography{references}  

\newpage
\section{Appendix}
\label{sec:appendix}

\subsection{Training Details}
\subsubsection{Observation Space}
The observation vector $\bm{s}_t \in \mathbb{R}^{187}$ consists of five main components: joint position history, proprioceptive feedback, privileged information, previous actions, and command signals. Table~\ref{tab:observation_space} summarizes each observation component.
\begin{table}[!tbh]
\centering
  \begin{tabular}{l l l l}
    \toprule
    \textbf{Observation Group} & \textbf{Observation Name} & \textbf{Dimension} & \textbf{Group Dim} \\ \midrule
    History $\bm{h}_t$ & Joint position history & 72 & 72 \\ \midrule
    \multirow{5}{*}{Proprioception $\bm{o}_t$} & Projected gravity & 3 & \multirow{5}{*}{45} \\ 
    & Base linear velocity & 3 &  \\
    & Base angular velocity & 3 &  \\ 
    & Joint position & 18 &  \\ 
    & Joint velocity & 18 &  \\ \midrule
    \multirow{8}{*}{Privileged info $\bm{p}_t$} & Feet contact state & 4 & \multirow{8}{*}{32} \\ 
    & Static friction & 4 &  \\ 
    & Feet air time & 4 &  \\ 
    & Base external wrench & 6 &  \\ 
    & Base external push velocity & 6 &  \\ 
    & Base mass disturbance & 1 &  \\ 
    & End-effector external wrench & 6 &  \\ 
    & End-effector mass disturbance & 1 &  \\ \midrule
      Previous action $\bm{a}_{t-1}$  & Previous actions & 18 & 18 \\ \midrule
    \multirow{4}{*}{Commands $\bm{c}_t$} & Desired Base Velocity & 3 & \multirow{4}{*}{20} \\ 
    & Desired End-Effector twist command & 6 &  \\
    & Desired End-Effector final goal pose & 7 & \\
    & Desired Feet swing heights & 4 &  \\ \bottomrule \\ 
  \end{tabular}
  \caption{Observation space for the teacher policy.}
  \label{tab:observation_space}
\end{table}
\begin{table}[!tbh]
\centering
  \begin{tabular}{l l l}
    \toprule
    \textbf{Observation Group} & \textbf{Observation Name} & \textbf{Noise range} \\ \midrule
    History & Joint position history & $[-0.01, 0.01]$ \\ \midrule
    \multirow{5}{*}{Proprioception} & Projected gravity & $[-0.01, 0.01]$ \\ 
    & Base linear velocity & $[-0.01, 0.01]$ \\ 
    & Base angular velocity & $[-0.1, 0.1]$ \\ 
    & Joint position & $[-0.01, 0.01]$ \\ 
    & Joint velocity & $[-0.2, 0.2]$ \\ \midrule
    Previous actions & Previous actions & - \\ \midrule
    \multirow{3}{*}{Commands} & Desired Base Velocity & - \\ 
    & Desired EE twist & - \\ 
    & Desired Feet swing heights & - \\
    & Desired Feet swing heights & - \\ \bottomrule \\
  \end{tabular}
  \caption{Observation space for the student policy and the range of noise applied during teacher-student distillation.}
  \label{tab:observation_noise}
\end{table}
The student policy is trained without any privileged information such that it can be deployed on hardware. During teacher-student distillation, uniform noise is added to the observations that are passed to the student policy. Table~\ref{tab:observation_noise} summarizes the noise added to each observation group.

\subsubsection{Rewards}
The rewards are categorized into three groups: locomotion, manipulation, and contact schedule. For most quantities, we employ a Gaussian tracking reward function $\Phi(\bm{v}, \sigma^2) = \exp(-\bm{v}^T\bm{v}/\sigma^2)$. Table~\ref{tab:reward_function} summarizes the reward functions, while Table~\ref{tab:symbols} describes the symbols used in these functions. We observe that multi-critic learning requires significantly less reward tuning compared to single-critic approaches. Unlike previous works \cite{fu2022deepwholebodycontrollearning, zargarbashi2024robotkeyframing}, our method avoids weighted averaging during advantage estimation, which further reduces the number of hyperparameters that require tuning.
\begin{table}[!tbh]
\centering
  \begin{tabular}{l l l l}
    \toprule
    \textbf{Group} & \textbf{Reward Name} & \textbf{Reward Function} & \textbf{Weight} \\ \midrule
    \multirow{12}{*}{Loco} & Base linear velocity & $\Phi(\hat{\bm{v}}_{b_{x, y}} - \bm{v}_{b_{x, y}}, 0.1) $ & 2.0 \\
    & Base angular velocity & $\Phi(\hat{\bm{\omega}}_{b_z} - \bm{\omega}_{b_z}, 0.05) $ & 2.0 \\
    & Torso height & $\Phi(\hat{\bm{h}}_{b_{z}} - \bm{h}_{b_{z}}, 0.1) $ & 0.5 \\ 
    & Base roll pitch angles & $\Phi(\bm{\theta}_{b_{xy}}, 0.1) $ & 0.1 \\ 
    & Torso linear velocity & $\Phi(\bm{v}_{b_{z}}, 0.2) $ & 0.5 \\ 
    & Torso roll pitch velocities & $\Phi(\bm{\omega}_{b_{xy}}, 0.2) $ & 2.5 \\ 
    & Is alive & $!z_{terminated}$ & 0.05 \\ 
    & Is terminated & $z_{terminated}$ & -400.0 \\ 
    & Undesired robot contacts & $n_{contacts, robot}$ & -1.0 \\ 
    & Robot action rate & $\Phi(\bm{a}_{t, robot} - \bm{a}_{t-1, robot}, 0.1)$ & $0.001$ \\ 
    & Robot joint torque &  $\Phi(\bm{\tau}_{t, robot}, 40.0)$ & $0.00001$ \\ 
    & Robot joint velocity &  $\Phi(\dot{\bm{q}}_{t, robot}, 4.0)$ & $0.0001$ \\ \midrule
    \multirow{6}{*}{Mani} & End-Effector position & $\Phi(r_{{EE}_t} - (r_{{EE}_{t-1}} + \hat{v}_{EE} \cdot \Delta t), 0.005) $ & 5.0 \\
    & End-Effector orientation & $\Phi(\bm{R}_{{EE}_t} \boxminus (\bm{R}_{{EE}_{t-1}} \boxplus \hat{\omega}_{EE} \cdot \Delta t), 0.01) $ & 4.0 \\
    & Undesired arm contacts & $n_{contacts, arm}$ & -1.0 \\ 
     & Arm action rate & $\Phi(\bm{a}_{t, robot} - \bm{a}_{t-1, robot}, 0.5)$ & $0.1$ \\ 
    & Arm joint torque &  $\Phi(\bm{\tau}_{t, robot}, 40.0)$ & $0.00001$ \\ 
    & Arm joint velocity &  $\Phi(\dot{\bm{q}}_{t, robot}, 4.0)$ & $0.0001$ \\ \midrule
    \multirow{4}{*}{CS} & Feet Contact & \begin{small}$\Sigma_f (1 - C_f)\cdot \Phi(F_f, 1.0)\cdot \Phi(\hat{h_z}-h_z, 0.05) + $\end{small} & 1.0 \\
    &  & \begin{small}$\Sigma_f C_f\cdot n_{F_{f_z} > 1.0}\cdot \Phi(v_{f_{xy}}, 0.01)$\end{small} & \\
    & Feet air time variance & $\text{Var}(t_{air_{t-2..t}}) + \text{Var}(t_{contact_{t-2..t}})$ & 1.0 \\
    & Feet air time & $\Sigma_{i=1}^{4} n_{contacts, feet}\cdot t_{air_i}$ & 0.25 \\ \bottomrule \\
  \end{tabular}
  \caption{Reward functions for training the teacher policy categorized into three groups, Loco (locomotion), Mani(manipulation and CS (contact schedule).}
  \label{tab:reward_function}
\end{table}

\begin{table}[!tbh]
\centering
  \begin{tabular}{l l}
    \toprule
    \textbf{Symbol} & \textbf{Description} \\ \midrule
    $\hat{\bm{v}}_{b_{x, y}}$ & Desired base linear velocity in x and y direction \\
    $\hat{\bm{\omega}}_{b_z}$ & Desired base angular velocity in z direction \\
    $\hat{\bm{h}}_{b_{z}}$ & Desired torso height \\
    $\bm{\theta}_{b_{xy}}$ & Base roll and pitch angles \\
    $\bm{v}_{b_{z}}$ & Torso linear velocity in z direction \\
    $\bm{\omega}_{b_{xy}}$ & Torso roll and pitch velocities \\
    $z_{terminated}$ & Termination signal \\
    $n_{contacts, robot}$ & Number of undesired robot contacts \\
    $n_{contacts, arm}$ & Number of undesired arm contacts \\
    $\bm{a}_{t, robot}$ & Robot action at time t \\
    $\bm{\tau}_{t, robot}$ & Robot joint torques at time t \\
    $\dot{\bm{q}}_{t, robot}$ & Robot joint velocities at time t \\
    $r_{{EE}_t}$ & End-effector position at time t \\
    $\bm{R}_{{EE}_t}$ & End-effector orientation at time t \\
    $\hat{v}_{EE}$ & Desired end-effector velocity \\
    $\hat{\omega}_{EE}$ & End-effector angular velocity \\
    $C_f$ & Feet contact probability \\
    $F_f$ & Feet contact force \\
    $h_z$ & Feet height \\
    $v_{f_{xy}}$ & Feet velocity in x and y direction \\
    $t_{air_i}$ & Feet air time \\
    $t_{contact_i}$ & Feet contact time \\
    $n_{F_{f_z} > 1.0}$ & Number of feet force signal in z direction \\ \bottomrule \\
  \end{tabular}
  \caption{Description of symbols used in the reward function.}
  \label{tab:symbols}
\end{table}

\subsubsection{Training Setup}
We train the policy by simulating 4096 parallel agents using IsaacLab \cite{mittal2023orbit}, a GPU-accelerated simulation framework. The training process incorporates a curriculum approach where agents begin on flat terrain and progressively advance to more challenging, moderately rough terrains as their performance improves. To facilitate effective zero-shot sim-to-real transfer, we implement extensive domain randomization across multiple physical parameters, as comprehensively documented in Table~\ref{tab:disturbances}.
\begin{table}[!tbh]
\centering
  \begin{tabular}{l l}
    \toprule
    \textbf{Disturbance} & \textbf{Range} \\ \midrule
    Feet static friction & $[0.5, 1.2]$ N \\
    Feet dynamic friction & $[0.3, 1.2]$ N \\
    Torso mass & $[-10, 10]$ kg \\
    End-effector mass & $[0, 1.8]$ kg \\
    External force on torso & $[-50, 50]$ N \\
    External torque on torso & $[-20, 20]$ Nm \\
    External force on end-effector & $[-3, 3]$ N \\
    Torso push linear velocity & $[-0.2, 0.2]$ m/s \\
    Torso push angular velocity & $[-0.2, 0.2]$ rad/s \\ \bottomrule \\
  \end{tabular}
  \caption{Disturbances applied to the robot during training.}
  \label{tab:disturbances}
\end{table}
We observe that randomization of foot friction and the application of external disturbances to the torso contribute to the development of stable walking behaviors and enhance the robustness of the policy during rapid arm movements. The external torso push velocity components are implemented by augmenting the measured torso velocity with random values for variable durations during simulation. This perturbation strategy teaches the policy to maintain and recover balance when faced with unexpected environmental disturbances.

The teacher policy architecture comprises an actor MLP network with 3 hidden layers ($[512, 256, 128]$ units) and ReLU activations. The multi-critic implementation consists of 3 critic MLP networks with identical dimensions. The policy is trained using PPO \cite{schulman2017proximal}. During both teacher and student policy trainings, the control policy operates at 50 Hz. While the control policy runs at an 8:1 decimation ratio, the simulation itself runs at 400 Hz to accurately model the robot's main control frequency. We list the hyperparameters used for training in Table \ref{tab:hyperparameters}.

\begin{table}[!htb]
\centering
\begin{minipage}{0.45\textwidth}
\centering
\resizebox{\textwidth}{!}{
  \begin{tabular}{l l}
    \toprule
    \textbf{Hyperparameter} & \textbf{Value} \\ \midrule
    Learning rate & $3.0e-4$ \\
    Entropy coefficient & $0.002$ \\
    Value loss coefficient & $1.0$ \\
    Clip parameter & $0.2$ \\
    Number of learning epochs & $8$ \\
    Number of mini-batches & $4$ \\
    Discount factor $\gamma$ & $0.99$ \\
    GAE $\lambda$ & $0.95$ \\
    Number of steps per environment & $24$ \\
    Number of environments & $4096$ \\
    Number of training iterations & $50000$ \\ \bottomrule \\
  \end{tabular}
  }
  \caption{Hyperparameters used for training the teacher policy.}
  \label{tab:hyperparameters}
  \end{minipage}
  \hfill
  \begin{minipage}{0.45\textwidth}
\centering
\resizebox{\textwidth}{!}{
\begin{tabular}{l l}
    \toprule
    \textbf{Hyperparameter} & \textbf{Value} \\ \midrule
    Learning rate & $1.0e-3$ \\
    Number of learning epochs & $8$ \\
    Gradient length & $15$ \\ 
    Loss type & Mean Squared Error \\ \bottomrule \\
  \end{tabular}
}
  \caption{Hyperparameters used for training the student policy.}
  \label{tab:student_hyperparameters}
\end{minipage}
\end{table}

\subsubsection{Teacher-Student Distillation}
Our teacher policy is a 3-layer MLP with hidden layers of size $[512, 256, 128]$ with ELU activations. The teacher has access to privileged information inaccessible in the real world. To deploy the policy on hardware, a student policy with the same network structure as the teacher is distilled without access to privileged information using supervised learning~\cite{lee2020learning}. During distillation, the student learns to implicitly estimate the privileged information by minimizing the mean squared error between the teacher and student actions. The hyperparameters used for training the student policy are listed in Table \ref{tab:student_hyperparameters}.

\subsection{Ablation Studies}

\subsubsection{Effect of Curriculum Learning on End-Effector Twist Commands}
\begin{figure}[!tbh]
    \centering
    \includegraphics[width=1.0\linewidth]{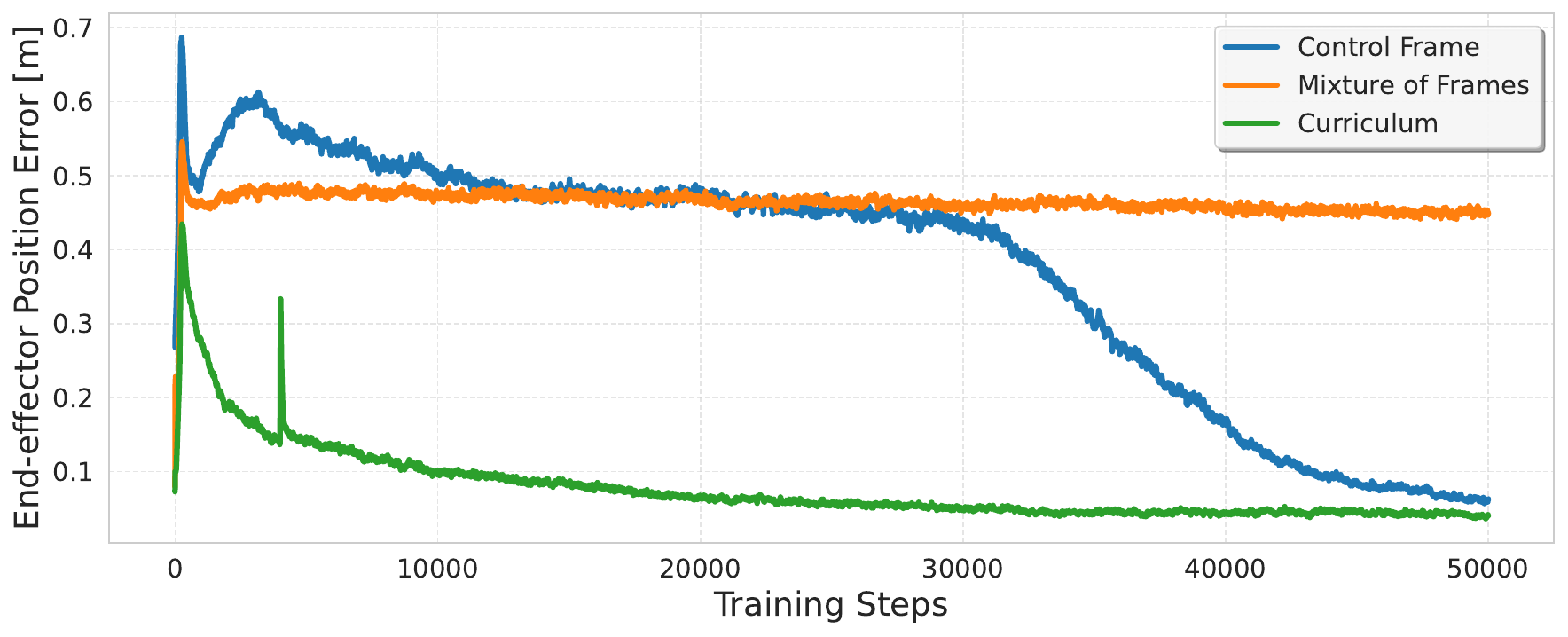}
    \caption{\textbf{End-effector position tracking for different command frame definitions}: The curriculum approach demonstrates superior performance compared to using either fixed frame representation or a mixture of frames.}
    \label{fig:command_frame}
\end{figure}

Similar to previous works \cite{fu2022deepwholebodycontrollearning, portela2024learningforcecontrollegged} in pose-based end-effector control, we formulate the end-effector twist commands in the control frame, which is defined as the robot-centric gravity-aligned frame that follows only the yaw of the robot's torso. However, as shown by the Control Frame signal in Figure \ref{fig:command_frame}, representing the twist command in the control frame makes it challenging for the policy to learn to precisely follow the end-effector trajectory. 
We observe that as the robot learns to stabilize its torso and walk, the control frame moves significantly, making the state-reward pairs sparse in representation and leading to difficulties in policy learning. As a solution, we attempted two approaches:
\begin{itemize}
\setlength\itemsep{-0.2em}
    \item \textbf{Mixture of frames}: In this approach, we provide 50\% of the commands to the policy in the base frame, while representing the rest in the control frame.
    \item \textbf{Curriculum}: In this approach, we represent the commands in the base frame until a set number of iterations, after which the commands are provided with respect to the control frame.
\end{itemize}
In Figure \ref{fig:command_frame}, it can be seen that the mixture of frames does not lead to better end-effector position tracking. This could be due to the conflicting representations caused by the two frames in which the commands are represented. In contrast, we observe that the curriculum approach leads to faster policy learning. In this case, we command the end-effector twist command with respect to the base frame until 3000 iterations, following which the frame switches to the control frame. The switching iteration is based on the observation that robots learn to walk with a stable torso within 3000 iterations. Although we see a jump in the tracking error soon after the frame switch, the tracking error continues to reduce for the rest of the training.

\subsubsection{Single-Critic vs Multi-Critic Learning}
\begin{figure}[!tbh]
    \centering
    \includegraphics[width=1.0\linewidth]{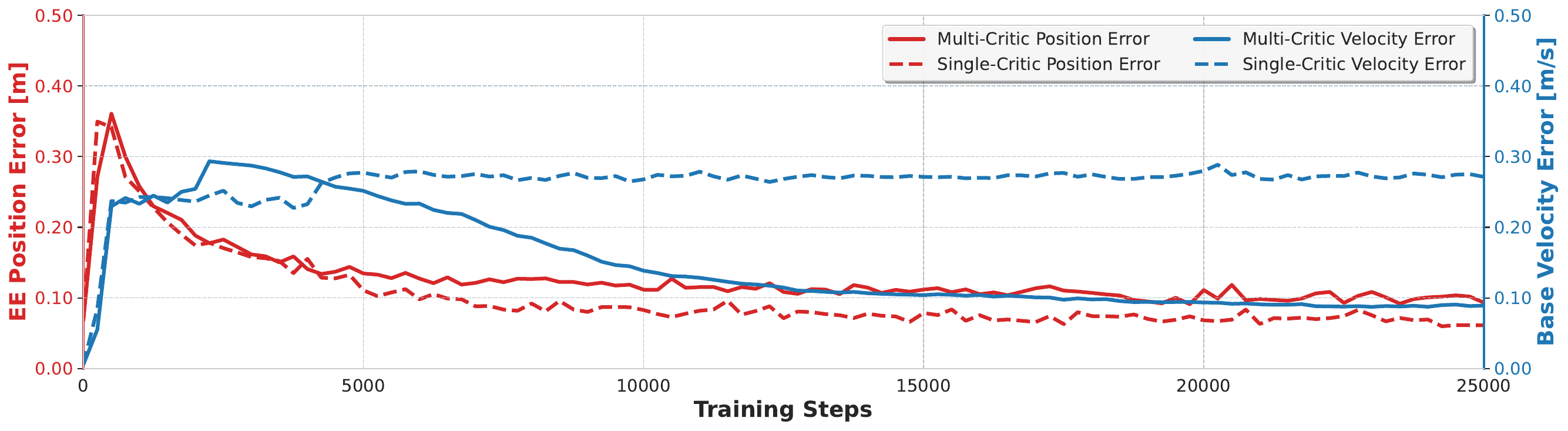}
    \caption{\textbf{Comparison of single-critic vs multi-critic frameworks}: Multi-critic policy learns to track both the base and end-effector commands whereas the single-critic policy only learns to track the end-effector commands.}
    \label{fig:single_critic_multi_critic}
\end{figure}

In this section, we compare the performance of single-critic versus multi-critic learning approaches. We train policies with identical parameters, modifying only the critic architecture, while maintaining consistent reward functions and weights. As shown in Fig.~\ref{fig:single_critic_multi_critic}, we observe an interesting trade-off: the single-critic policy achieves marginally lower end-effector tracking errors, but completely fails to learn locomotion behaviors.
This superior end-effector performance can be attributed to the policy that adopts a stationary posture at all times that essentially ignores base-velocity commands, allowing it to focus on the arm control task. In contrast, the multi-critic approach successfully learns to simultaneously satisfy both locomotion and manipulation objectives, demonstrating the ability to coordinate whole-body movement while maintaining precise end-effector motion.

\subsubsection{Chicken head control performance while walking}

\begin{wrapfigure}{r}{0.6\textwidth}
\vspace{-4mm}
  \begin{center}
    \includegraphics[width=1.0\linewidth]{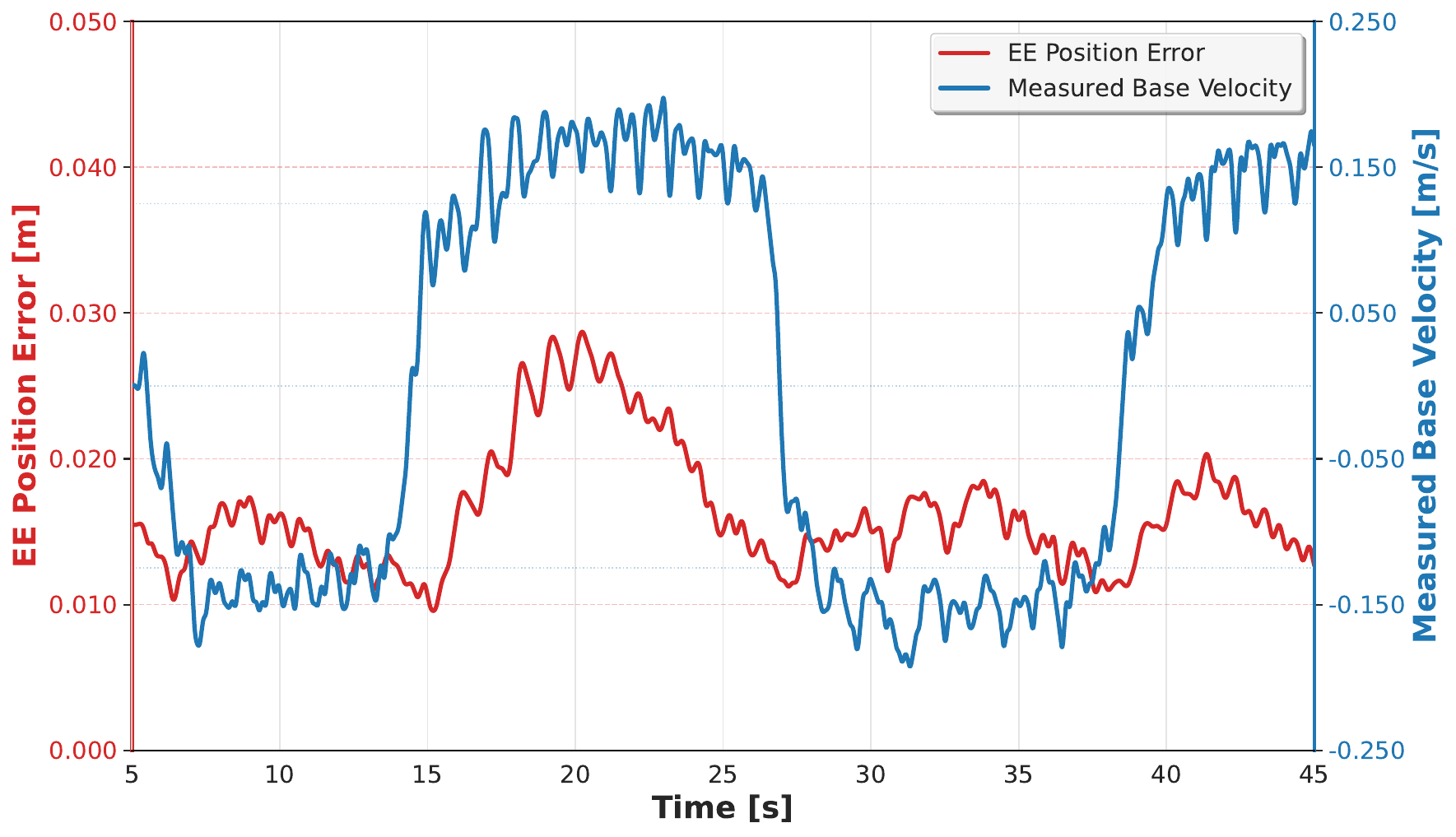}
    \end{center}
    \caption{\textbf{End-Effector Tracking performance in chicken-head mode}: Tracking error in end effector position (red, left axis) and measured base velocity (blue, right axis) of the robot plotted over time. The position error has a mean of 0.0156 m and maximum of 0.0287 m.}
    \label{fig:3d_traj_chicken_mode}
  \vspace{-5mm}
  \label{fig:perf_over_forces}
\end{wrapfigure}

To demonstrate the whole-body task coordination capabilities of the controller, we perform chicken head control with the end-effector while walking. The goal of this experiment is to demonstrate the whole-body loco-manipulation capability of the policy by holding a static end-effector pose in the world frame while moving the torso. During this experiment, the robot is commanded a base velocity within the range of [-0.2, 0.2] m/s while the end-effector is commanded twist corrections to hold the end-effector position in the robot's odometry frame. As shown in Fig. \ref{fig:3d_traj_chicken_mode}, the policy achieves very low end-effector position tracking errors while walking, demonstrating effective task space control during base motion.

\end{document}